\documentclass[letterpaper]{article} 
\usepackage[]{aaai24}  
\usepackage{times}  
\usepackage{helvet}  
\usepackage{courier}  
\usepackage[hyphens]{url}  
\usepackage{graphicx} 
\usepackage{natbib}  
\usepackage{caption} 
\usepackage{algorithm}
\usepackage{algorithmic}

\usepackage{newfloat}
\usepackage{listings}
\DeclareCaptionStyle{ruled}{labelfont=normalfont,labelsep=colon,strut=off} 
\floatstyle{ruled}
\newfloat{listing}{tb}{lst}{}
\floatname{listing}{Listing}
\nocopyright 

\title{\texttt{appjsonify}: An Academic Paper PDF-to-JSON Conversion Toolkit}

\author{
    Atsuki Yamaguchi\textsuperscript{\rm 1}\thanks{Work done while at Hitachi, Ltd.} {\rm and} Terufumi Morishita\textsuperscript{\rm 2}
}
\affiliations {
    \textsuperscript{\rm 1}Department of Computer Science, University of Sheffield, United Kingdom\\
    \textsuperscript{\rm 2}Research and Development Group, Hitachi, Ltd., Japan\\
    ayamaguchi1@sheffield.ac.uk, terufumi.morishita.wp@hitachi.com
}

\begin{document}
\maketitle

\begin{abstract}
We present \texttt{appjsonify}\footnote{Named after \textbf{A}cademic \textbf{P}aper \textbf{P}DF \textbf{jsonify}.}, a Python-based PDF-to-JSON conversion toolkit for academic papers.
It parses a PDF file using several visual-based document layout analysis models and rule-based text processing approaches.
\texttt{appjsonify} is a flexible tool that allows users to easily configure the processing pipeline to handle a specific format of a paper they wish to process.
We are publicly releasing \texttt{appjsonify} as an easy-to-install toolkit available via PyPI\footnote{\url{https://pypi.org/project/appjsonify/}} and GitHub\footnote{\url{https://github.com/hitachi-nlp/appjsonify}}.
\end{abstract}

\section{Introduction}
A tremendous amount of academic papers have been published every year, which are typically distributed in Portable Document Format (PDF).
Obtaining clean and structured text from them can be very useful for research and development activities in the field of artificial intelligence (AI), from foundational model training~~\cite{beltagy-etal-2019-scibert} to machine learning (ML)-based applications such as paper recommendation~\cite{Kreutz2022} and fact-checking using papers~\cite{tan-etal-2023-multi2claim}.

Document layout analysis (DLA) is one of the main tools to obtain clean and structured text from scientific documents~\cite{shen-etal-2022-vila}, and several datasets related to scientific domains have been proposed.
For instance, TableBank~\cite{li-etal-2020-tablebank} provides an image-based table detection and recognition dataset, while PubLayNet~\cite{8977963} offers token-level annotations, whose category includes text, title, list, table, and figure.
DocBank~\cite{li-etal-2020-docbank} has more fine-grained annotations than PubLayNet and covers 13 categories.
In addition to the datasets, many DLA methods have also been developed for structured scientific content extraction, such as LayoutLM~\cite{10.1145/3394486.3403172} and VILA~\cite{shen-etal-2022-vila}.

Despite the many advances in DLA, it is still challenging to structure scientific documents using a \textit{single} DLA model.
This is because papers often have variations in their formats, such as the number of columns, citation style, fonts, and the position of captions.
The single is not robust enough to cope with these many format differences.

\begin{figure*}[t]
\centering
\includegraphics[width=\textwidth]{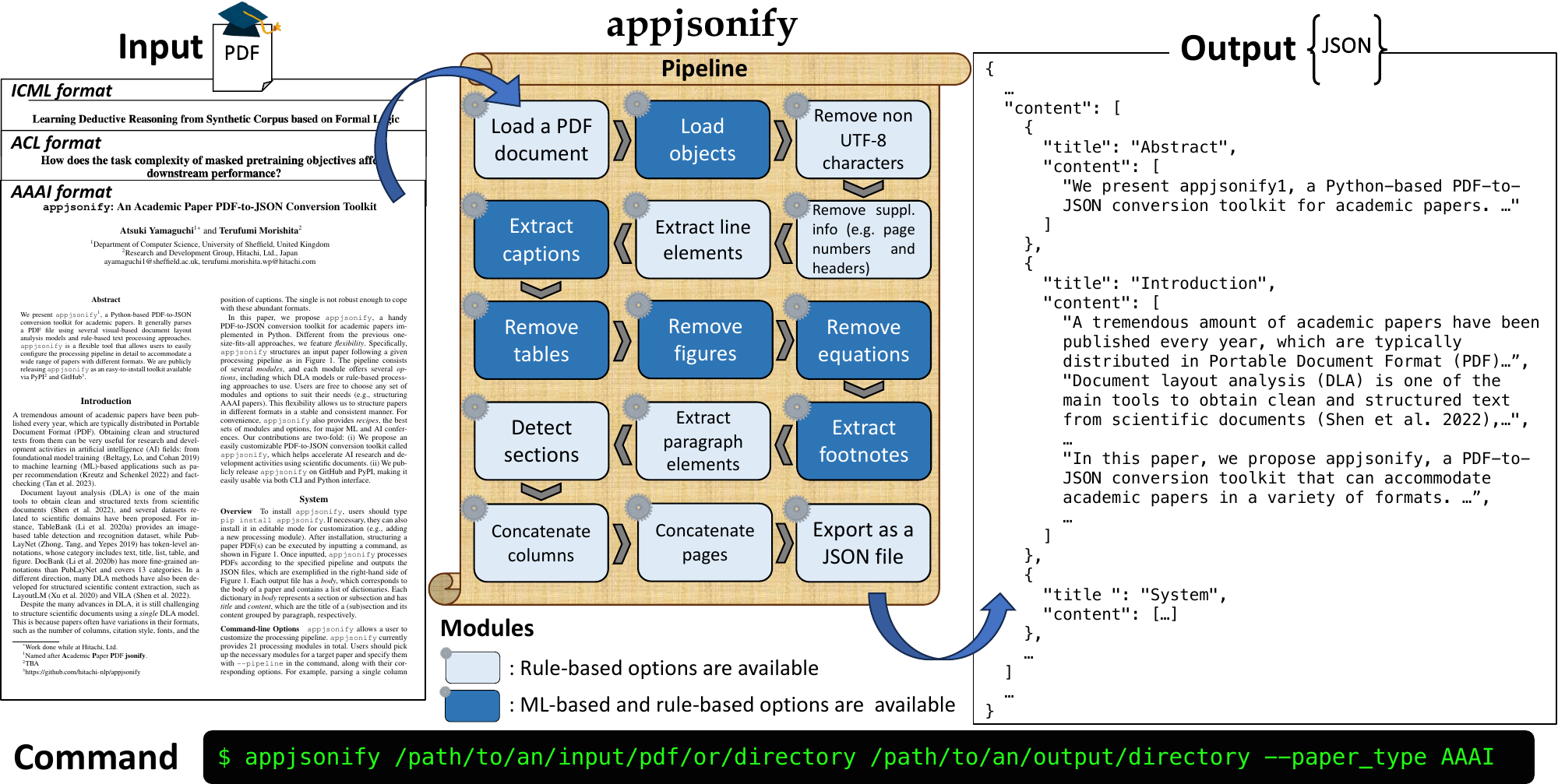} 
\caption{
An overview of \texttt{appjsonify}, which converts a paper PDF file into a JSON file grouped by paragraph.
The pipeline will be automatically configured by specifying \texttt{paper\_type} if a target paper is from major AI and ML conferences.
}
\label{fig1}
\end{figure*}

In this paper, we propose \texttt{appjsonify}, a PDF-to-JSON conversion toolkit that can accommodate academic papers in a variety of formats.
To handle these, \texttt{appjsonify} structures an input paper following a given processing pipeline as in Figure \ref{fig1}.
The pipeline consists of several \textit{modules}, and each module offers several \textit{options}, including which DLA models or rule-based processing approaches to use.
Users are free to choose any set of modules and options that best fit the specific paper format (e.g., AAAI format) they are targeting, unlike the previous one-size-fits-all approaches.

For convenience, \texttt{appjsonify} also provides \textit{recipes}, the best sets of modules and options, for major ML and AI conferences such as AAAI, ICML, NeurIPS, and ACL.
We publicly release \texttt{appjsonify} on GitHub and PyPI, which should help accelerate AI research and development activities using scientific documents.
Finally, \texttt{appjsonify} provides both CLI and Python interface, making it easily usable.

\section{System}
\paragraph{Overview}
To install \texttt{appjsonify}, users should type \texttt{pip install appjsonify}.
If necessary, they can also install it in editable mode for customization (e.g., adding a new processing module).
After installation, structuring a paper PDF(s) can be executed by inputting a command, as shown in Figure \ref{fig1}.
Once inputted, \texttt{appjsonify} processes PDFs according to the pipeline and outputs the JSON files, which are exemplified in the right-hand side of Figure \ref{fig1}.
Each output file has a \textit{body}, which corresponds to the body of a paper and contains a list of dictionaries.
Each dictionary in \textit{body} represents a section or subsection and has \textit{title} and \textit{content}, which are the title of a (sub)section and its content grouped by paragraph, respectively.

\paragraph{Command-line Options}
In \texttt{appjsonify}, users can design their own PDF extractor tailored to their needs by customizing the processing pipeline.
They should pick up the necessary modules for a target paper(s) and specify the modules with \texttt{--pipeline} in the command, along with their corresponding options.
For example, parsing a single column paper like \citet{li-etal-2020-docbank} can be executed with the following command:
\begin{quote}
\begin{scriptsize}
\begin{verbatim}
appjsonify /path/to/an/input/pdf/or/directory \
/path/to/an/output/directory \
    --pipeline load_docs load_objects_with_ml \
    remove_illegal_tokens remove_meta extract_lines \
    extract_captions_with_ml remove_figures_with_ml \
    remove_tables_with_ml remove_equations_with_ml \
    extract_footnotes_with_ml extract_paragraphs \
    detect_sections concat_pages dump_formatted_doc \
    --preset_table_caption_pos below \
    --preset_figure_caption_pos below \
    --consider_font_size \
    --headline_names Abstract
\end{verbatim}
\end{scriptsize}
\end{quote}
\noindent Each module in the command is specified with a whitespace delimiter, followed by the options.
The selected modules are almost identical to those in Figure \ref{fig1}, except for the removal of ``Concatenate columns'', since no concatenation is needed for the single-column paper.
For the options, the first and second options are to help match captions to the correct figure or table.
The third option lets \texttt{appjsonify} consider font size when merging objects in \texttt{extract\_pragraphs}, \texttt{concat\_pages}, and \texttt{detect\_sections}.
The last option helps \texttt{appjsonify} recognize unnumbered (sub)section names in \texttt{detect\_sections} and properly structure the contents.

\paragraph{Implementations}
\texttt{appjsonify} mainly uses the following publicly available resources.
\begin{enumerate}
    \item \textbf{Pdfplumber}~\cite{Singer-Vine_pdfplumber_2023} loads a PDF file and returns raw texts with their bounding box and font information, as well as object bounding boxes such as rectangles, lines, curves, and images.
    
    \item \textbf{Faster R-CNN}~\cite{NIPS2015_14bfa6bb} serves as a bounding box detector for objects.\footnote{Objects include a table, figure, caption, and equation.}
    
    \item \textbf{DocBank}, \textbf{PubLayNet}, and \textbf{TableBank} were each used to train the bounding box detectors.
\end{enumerate}

\noindent Each module can use these as it needs.
For example, ``Load objects'' in Figure \ref{fig1} (\texttt{load\_objects\_with\_ml}) uses 2 and 3, and ``Detect sections'' (\texttt{detect\_sections}) uses font information obtained in 1.

\paragraph{Recipes}
For convenience, \texttt{appjsonify} provides recipes, which are the best sets of modules and options for major AI and ML conferences.
Currently, we support AAAI, ACL, ICML, ICLR, NeurIPS, ACM, IEEE, and Springer papers.
To use the recipe for AAAI, for instance, all we have to do is specify the example command shown in Figure \ref{fig1}.
We will then obtain its parsed JSON file as exemplified in Figure \ref{fig1}.

\section{Conclusion}
We proposed \texttt{appjsonify}, a handy academic paper PDF to JSON conversion tool available via PyPI.
\texttt{appjsonify} is a versatile tool in that it can handle various academic papers in different formats, thanks to the use of different DLA models and rule-based approaches and the easily customizable approach.
For future releases, we plan to incorporate new DLA models (e.g., Nougat~\cite{blecher2023nougat}) into \texttt{appjsonify} to further improve robustness and support multilingual documents other than English.

\clearpage
\bibliography{main,anthology}
\end{document}